\documentclass{article}
\usepackage{fix-cm}

\PassOptionsToPackage{table}{xcolor}
\usepackage{iclr2027_conference,times}
\iclrfinalcopy
\usepackage[utf8]{inputenc}
\usepackage[T1]{fontenc}
\usepackage{url}
\usepackage{booktabs}
\usepackage{amsfonts}
\usepackage{amsmath}
\usepackage{amssymb}
\usepackage{nicefrac}
\usepackage{microtype}
\usepackage[table]{xcolor}
\usepackage{graphicx}
\usepackage{float}
\usepackage{placeins}

\usepackage{multirow}
\usepackage{array}
\usepackage{tikz}
\usetikzlibrary{arrows.meta,svg.path}
\definecolor{oursbg}{RGB}{250,240,232}
\definecolor{gaincolor}{RGB}{39,95,88}

\newcommand{\globeicon}{%
  \begin{tikzpicture}[x=4.4pt, y=4.4pt, baseline=-3.2pt, line width=0.65pt, draw=codeblue]
    \draw (0,0) circle (1);
    \draw (0,0) ellipse (0.48 and 1);
    \draw (-0.95, 0.32) -- (0.95, 0.32);
    \draw (-0.95, -0.32) -- (0.95, -0.32);
    \draw (0, -1) -- (0, 1);
  \end{tikzpicture}%
}

\newcommand{\githubicon}{%
  \begin{tikzpicture}[baseline=1.4pt]
    \fill[black!85] svg{M 4.600 9.200 C 2.058 9.200 0.000 7.141 0.000 4.600 C 0.000 2.564 1.317 0.845 3.145 0.236 C 3.375 0.195 3.461 0.333 3.461 0.454 C 3.461 0.564 3.456 0.926 3.456 1.311 C 2.300 1.098 2.001 1.593 1.909 1.851 C 1.857 1.984 1.633 2.392 1.438 2.501 C 1.276 2.587 1.046 2.800 1.432 2.806 C 1.794 2.812 2.053 2.472 2.139 2.334 C 2.553 1.639 3.214 1.834 3.479 1.955 C 3.519 2.254 3.640 2.455 3.772 2.570 C 2.748 2.685 1.679 3.082 1.679 4.841 C 1.679 5.342 1.857 5.756 2.151 6.078 C 2.104 6.193 1.944 6.664 2.196 7.297 C 2.196 7.297 2.582 7.417 3.461 6.825 C 3.829 6.929 4.220 6.981 4.611 6.981 C 5.002 6.981 5.393 6.929 5.761 6.825 C 6.641 7.423 7.026 7.297 7.026 7.297 C 7.279 6.664 7.118 6.193 7.072 6.078 C 7.366 5.756 7.544 5.347 7.544 4.841 C 7.544 3.076 6.469 2.685 5.445 2.570 C 5.612 2.426 5.756 2.151 5.756 1.719 C 5.756 1.104 5.750 0.609 5.750 0.454 C 5.750 0.333 5.836 0.190 6.066 0.236 C 7.939 0.865 9.200 2.624 9.200 4.600 C 9.200 7.141 7.141 9.200 4.600 9.200 Z};
  \end{tikzpicture}%
}
\usepackage{listings}
\usepackage{wrapfig}
\usepackage{capt-of}
\usepackage[hidelinks]{hyperref}
\makeatletter
\@ifundefined{nolinenumbers}{}{}
\makeatother
\definecolor{tilgshade}{RGB}{244,244,244}

\definecolor{deepgreen}{RGB}{0,110,0}
\definecolor{deepred}{RGB}{150,30,30}
\definecolor{codebg}{RGB}{248,250,252}
\definecolor{codeborder}{RGB}{198,208,222}
\definecolor{codeblue}{RGB}{30,92,160}

\lstdefinestyle{tacgpseudocode}{
  language=Python,
  alsoletter={_},
  morekeywords={Tensor,Config,State,softmax,argmax,gather,detach,where,zeros_like,ones_like,clamp,masked_topk},
  basicstyle=\ttfamily\small,
  keywordstyle=\color{codeblue}\bfseries,
  emph={
    TACGCommitState,tacg_commit_step,z_ref,prev_proposal,proposal_streak,
    logits,mask,state,cfg,base_gate,p,x_hat,confidence,ref_logits,
    readout_scores,support_readout,p_ref,support,readiness,base_ready,extra_ready,
    commit_mask,write_tokens,m_base,m_extra,tau_escape,tau_floor,K_extra
  },
  emphstyle=\color{deepred},
  commentstyle=\color{deepgreen},
  backgroundcolor=\color{codebg},
  frame=single,
  rulecolor=\color{codeborder},
  framerule=0.4pt,
  framesep=5pt,
  xleftmargin=2pt,
  xrightmargin=2pt,
  columns=fullflexible,
  keepspaces=true,
  breaklines=true,
  breakatwhitespace=true,
  showstringspaces=false,
  tabsize=2,
  aboveskip=0.5\baselineskip,
  belowskip=0.5\baselineskip
}

\title{TACG: Trajectory-Aware Commit Gating for Diffusion Language Model Decoding}

\author{%
  Chengcheng Wang\textsuperscript{1*}
  \quad
  Tingzhang Luo\textsuperscript{2*}
  \quad
  Wenhao Li\textsuperscript{1}
  \quad
  Jianyuan Guo\textsuperscript{2}
  \quad
  Chang Xu\textsuperscript{1} \\
  \textsuperscript{1}University of Sydney
  \quad
  \textsuperscript{2}City University of Hong Kong \\
  {\small\texttt{cwan0785@uni.sydney.edu.au}
  \quad
  \texttt{tingzhluo3-c@my.cityu.edu.hk}} \\
  {\small\texttt{weli0055@uni.sydney.edu.au}
  \quad
  \texttt{jianyguo@cityu.edu.hk}
  \quad
  \texttt{c.xu@sydney.edu.au}}
}

\begin{document}

\maketitle
\ificlrfinal
\begingroup
\renewcommand{\thefootnote}{\fnsymbol{footnote}}
\footnotetext[1]{Equal contribution.}
\endgroup
\fi

\vspace{-1.2em}
\begin{center}
  \href{https://tacg.traly.org/}{\globeicon\hspace{0.4em}\textcolor{codeblue}{\textbf{Project Page}}}%
  \hspace{2.2em}%
  \href{https://github.com/Clarence-CV/TACG-DLLM}{\githubicon\hspace{0.4em}\textcolor{codeblue}{\textbf{GitHub}}}%
\end{center}
\vspace{0.2em}

\begin{abstract}
Diffusion language models predict many masked positions at each denoising step, but their efficiency and output quality depend on which predictions enter the evolving context. Confidence is informative but incomplete for this decision: at matched confidence, proposals with different identity and support histories exhibit different one-step revision rates. We propose Trajectory-Aware Commit Gating (TACG), a training-free decoder that separates the current token proposal from the decision to commit it. TACG scores temporal support for the current proposal by contrasting its logits with an exponential moving average of earlier logits, requires the proposal identity to persist through a History Gate, and promotes a capped number of additional positions per step. The gate decides which positions to commit and when; every written token remains the contemporaneous base proposal. On complete four-task LLaDA evaluations under the stated single-seed protocol, measured task-specific operating points reach 51.02\% mean accuracy with 67.82 mean model evaluations, compared with 48.00\% and 76.59 for Native Confidence. Under a matched per-event commit-count replay, history-guided selection exceeds confidence ranking by 8.3 accuracy points at identical per-sample NFE. History-reference interventions alter commitment timing while preserving the base-proposal write rule. Experiments on Dream and LLaDA2-Mini characterize quality--cost behavior across backbones and generation lengths.
\end{abstract}

\section{Introduction}
Diffusion language models (DLLMs) generate text by repeatedly predicting the values of masked positions~\citep{austin2023structureddenoisingdiffusionmodels,sahoo2024simpleeffectivemaskeddiffusion,nie2025largelanguagediffusionmodels}. A single forward pass proposes tokens at many positions, creating an opportunity for parallel generation. Realizing this opportunity requires a second decision: which proposals should enter the context now? In the non-revising decoders considered here, a committed token stays in the sequence and conditions subsequent predictions. Decoding therefore controls both the proposed token and when that proposal enters the sequence.

Confidence provides a natural rule for this decision, but a scalar threshold uses only the current probability. Raising it delays commitment; lowering it admits more candidates without distinguishing how their support developed. Each masked position also carries a history of proposed identities and logits across denoising steps. Two proposals with the same current confidence may therefore have different trajectories, such as repeated support for the same token or a recent switch to a new candidate. Recent stability-aware decoders such as KLASS exploit changes in the predictive distribution across steps~\citep{kim2026klassklguidedfastinference}, while the role of candidate-specific proposal history in the commitment decision remains open.

\begin{figure}[h]
  \centering
  \includegraphics[width=0.9\linewidth]{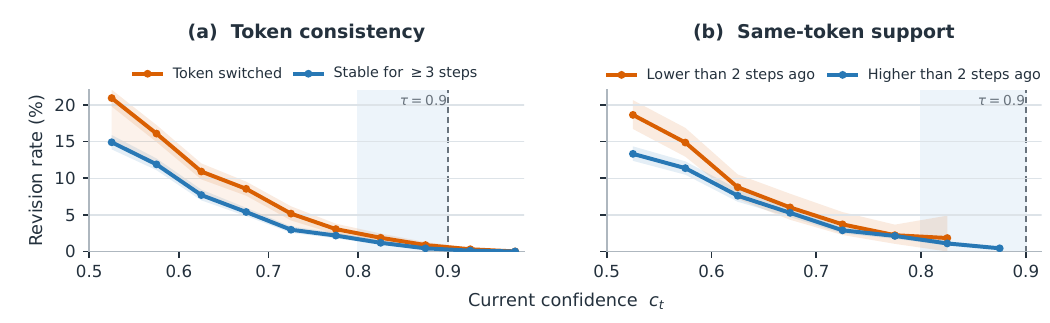}
  \caption{
  \textbf{Revision rates differ at matched confidence}. We delay commitment by one context update on GSM8K and record whether the model changes its top-1 proposal. (a) Within confidence bins, the model revises recently switched proposals more often than proposals that retain the same token for at least three steps. (b) After matching confidence and persistence, the model revises proposals with declining same-token support more often than proposals with increasing support.
  }
  \label{fig:support}
\end{figure}

We first test whether proposal persistence and support trend contain information that current confidence omits. A one-step hold (HOLD) probe keeps a candidate masked for one additional context update and records whether its top-1 identity changes. Figure~\ref{fig:support} reports two revision-rate differences on GSM8K. Within confidence bins, recently switched proposals revise more often than persistent proposals. After matching confidence and persistence, weakening proposals revise more often than strengthening proposals. The probe measures one-step proposal revision; it does not measure correctness or calibration.

This observation motivates a separation between proposing a token and committing its position. We introduce \emph{Trajectory-Aware Commit Gating} (TACG), a training-free decoder that uses two history signals. Figure~\ref{fig:tilg-mechanism} summarizes the resulting information flow. Temporal Implicit Logits Guidance (TILG) contrasts the current logits with an exponential-moving-average reference and scores the current proposal's support. A History Gate (HG) checks proposal persistence. TACG uses these signals to gate base commitments and rank a capped set of additional positions; every write remains the contemporaneous base proposal.

The separation defines a local write invariant: TACG changes which positions enter the context and when they enter it, while the token written at each accepted position comes from the base decoder at that step. Later context can therefore produce different subsequent proposals and different final text. We evaluate this mechanism with a selection-versus-writing intervention, a matched per-event commit-count replay, and logit-reference interventions.

Our headline LLaDA comparison uses complete HumanEval, MBPP, GSM8K, and MATH500 splits. The reported task-specific TACG operating points are measured on completed full-task runs from the candidate roster under the stated single-seed protocol; they add 3.02 accuracy points and reduce mean model calls by 11.45\% relative to Native Confidence. Dream and LLaDA2-Mini extend the analysis to a second backbone and longer generation lengths; we report matched timing and throughput studies with their own settings.

Our contributions are:
\begin{enumerate}
  \item We formulate parallel DLLM decoding as a commitment problem and introduce TACG, which combines candidate-conditioned temporal support with proposal persistence while retaining the base proposal at every write.
  \item We compare selection and writing with a 2$\times$2 intervention and test position choice with a matched per-event commit-count replay. The replay yields 49.42\% versus 41.16\% accuracy for history-guided versus confidence-ranked selection at identical per-sample NFE.
  \item We report complete quality--NFE results on three decoder families, together with logit-reference, sensitivity, timing, and length--batch studies that delimit how the tested gate changes decoding trajectories and cost.
\end{enumerate}

\section{Background: masked diffusion exposes a commitment decision}
\label{sec:background}
\subsection{The decoder chooses which proposals enter the context}
Given a masked sequence $x_t$, a DLLM predicts
$p_t(i,v)=p_\theta(x_0^i=v\mid x_t)$
over vocabulary entries $v$ at each masked position $i$.
A model call produces predictions for all active positions in parallel,
after which the decoder chooses a subset of proposals to reveal.
In the non-revising setting considered here, a revealed token remains in
the sequence and becomes context for subsequent model calls.
The model therefore supplies the current proposal, while the decoder
chooses when to commit that proposal.
LLaDA~\citep{nie2025largelanguagediffusionmodels} and
Dream~\citep{ye2025dream7bdiffusionlarge} instantiate this masked-diffusion
setting, while LLaDA2-Mini uses a block-diffusion variant~\citep{bie2025llada20scalingdiffusionlanguage}.

\subsection{A proposal trajectory contains discrete and continuous history}
Before the decoder commits a position, repeated denoising calls expose a trajectory
of its proposed identity and model support. For position $i$ at step $t$, let
$\hat{x}_{t,i}$ denote the current top-1 proposal and $c_{t,i}$ its confidence.
Proposal persistence counts consecutive observations that retain the same top-1 identity. For the current proposal $v=\hat{x}_{t,i}$ and $t\ge2$, a support-trend diagnostic is
\begin{equation}
  g_{t,i}
  =
  \log p_t(i,v)-\log p_{t-2}(i,v),
  \label{eq:diagnostic-trend}
\end{equation}
where positive values indicate increasing support for the current proposal and
negative values indicate decreasing support. This two-step diagnostic is separate
from the implemented EMA score $b_i$ in Section~\ref{sec:method}.

We use a one-step HOLD diagnostic to relate these histories to proposal
revision. If the decoder would commit position $i$ at step $t$, we keep it masked for
one additional model call while the decoder updates the surrounding context.
We record whether the top-1 proposal changes,
\begin{equation}
  R_{t,i}
  =
  \mathbb{1}\!\left[
  \hat{x}^{\mathrm{hold}}_{t+1,i}
  \neq
  \hat{x}_{t,i}
  \right].
  \label{eq:hold-revision}
\end{equation}
A revision $R_{t,i}=1$ means that the current proposal does not survive one
additional context update. It does not imply that either proposal is correct.
We use the HOLD probe only for analysis; it adds no model call during TACG
inference.

The comparisons in Figure~\ref{fig:support} stratify proposals by current
confidence before comparing recently switched and persistent proposals. They then
stratify by confidence and persistence before comparing decreasing and increasing
support.
These comparisons motivate a candidate-conditioned temporal signal, while the
method and experiments keep its interpretation operational: it ranks positions
for commitment and does not provide a correctness guarantee.

\section{Related Work}
\label{sec:related-work}
\noindent\textbf{Parallel commit policies.}
Fast-dLLM combines confidence-aware selection with approximate caching~\citep{wu2025fastdllmtrainingfreeaccelerationdiffusion}. APD adapts parallel sampling to the denoising state~\citep{israel2025acceleratingdiffusionllmsadaptive}, LocalLeap groups positions using local determinism~\citep{kong2025acceleratingdiffusionllminference}, Prophet uses a top-2 confidence gap to trigger early commits~\citep{li2026diffusionlanguagemodelsknow}, and Dynamic Sliding Block changes the reveal schedule with a dynamically sized sliding block~\citep{luo2026dsb}. These methods change the commit schedule through confidence, local consistency, or a predefined schedule. TACG uses the same commit interface and adds a candidate-specific temporal score.

\noindent\textbf{History and stability signals.}
KLASS combines confidence with a distribution-level KL signal~\citep{kim2026klassklguidedfastinference}. Anchor-based History-stable Decoding (AHD) uses historical anchors~\citep{zou2026ahd}, and DC-Leap uses draft-guided contiguous verification and leaping~\citep{jiao2026dcleap}. LookUM and LOPA evaluate candidate continuations or local lookahead~\citep{lee2025lookaheadunmaskingelicitsaccurate,xu2025lopascalingdllminference}. TACG reads logits already produced by the backbone, scores the current candidate relative to an exponential-moving-average reference, and introduces no auxiliary forward pass for the gate.

\noindent\textbf{Candidate-conditioned temporal contrast.}
TILG is algebraically analogous to classifier-free guidance because both form a contrastive readout from current and reference model outputs~\citep{ho2022classifierfreediffusionguidance}. Classifier-free guidance combines conditional and unconditional score estimates to form a sampling distribution. TILG contrasts current logits with earlier logits from the same model and uses the result to rank commitment candidates. The written token remains the base proposal, so TACG contributes a gate-level temporal readout alongside the base token distribution.

\section{Temporal history selects commit positions}
\label{sec:method}
\subsection{The base model supplies every written token}
\begin{figure}[htbp]
  \centering
  \includegraphics[width=\linewidth]{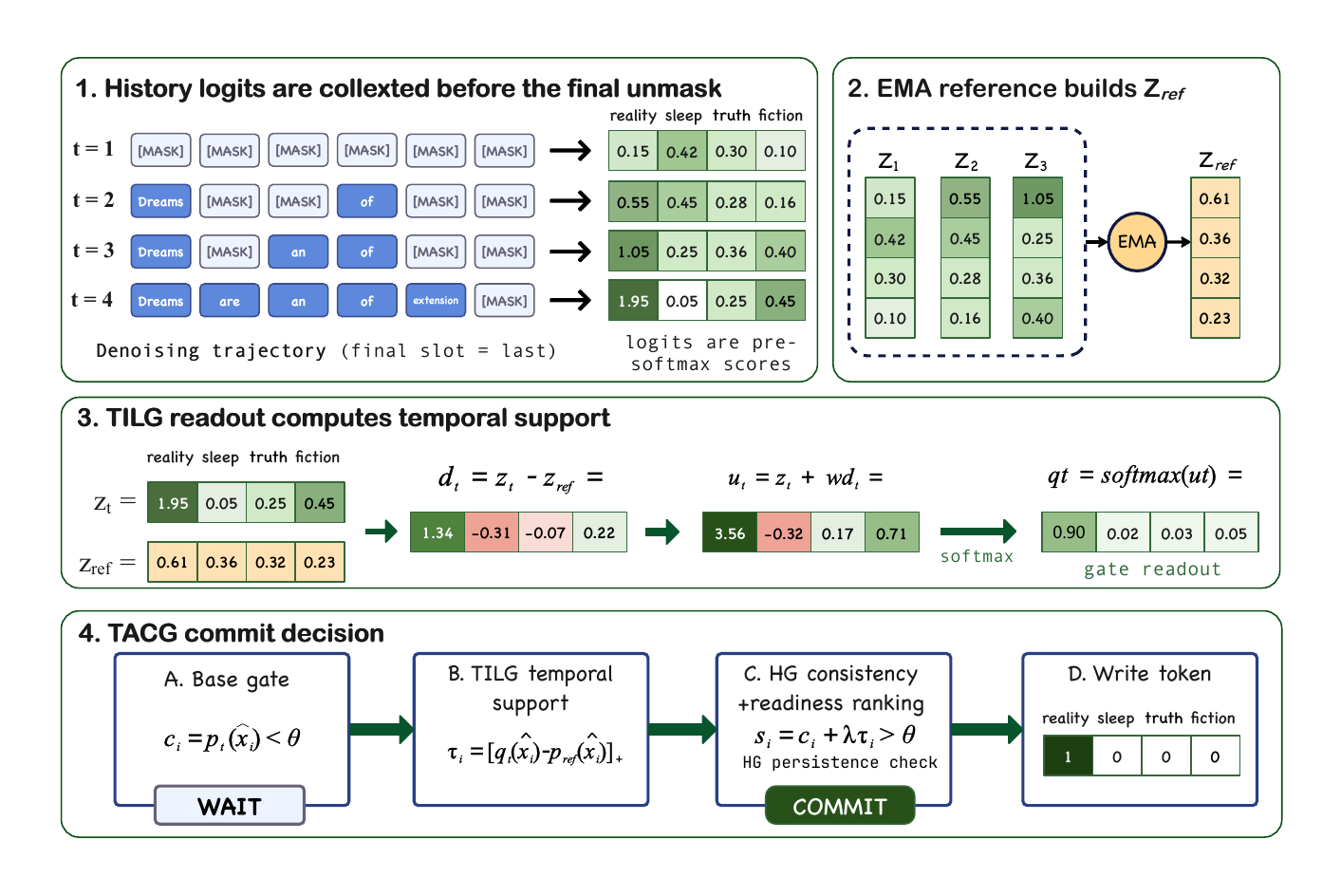}
  \caption{TACG uses a temporal-support branch to select commit positions. The base posterior supplies the current proposal. Earlier logits form an EMA reference, and TILG contrasts current logits with that reference to score candidate-conditioned support. History Gate and capped ranking select positions for commitment. Each write uses the current base proposal. The threshold sketch is illustrative; extra commitments follow the capped ranking rule in Section~\ref{sec:promotion}.}
  \label{fig:tilg-mechanism}
\end{figure}

Let $\mathcal M_t$ be the masked positions at step $t$, with logits $z_t(i,\cdot)$ and base posterior $p_t=\mathrm{softmax}(z_t)$. The base decoder supplies a proposal $\hat x_i$ and confidence $c_i=p_t(i,\hat x_i)$:
\begin{equation}
  \hat x_i=\arg\max_v p_t(i,v),\qquad c_i=p_t(i,\hat x_i).
  \label{eq:base-proposal}
\end{equation}
The gate decides whether to write this proposal. Figure~\ref{fig:tilg-mechanism} shows the information flow.

\subsection{TILG scores candidate-conditioned temporal support}
Current confidence describes a proposal's probability. Temporal support describes
how the model's relative evidence for that proposal has changed. We maintain an
EMA of earlier logits and read it before updating the reference:
\begin{equation}
  \bar z_t=\beta\bar z_{t-1}+(1-\beta)z_t.
  \label{eq:ema}
\end{equation}
The first observation initializes the reference with current logits, so its
support score is neutral. With $p_t^{\rm ref}=\mathrm{softmax}(\bar z_{t-1})$,
the TILG readout is
\begin{equation}
  q_t=\mathrm{softmax}\!\left(z_t+w(z_t-\bar z_{t-1})\right),
  \qquad q_t(v)\propto\frac{p_t(v)^{1+w}}{p_t^{\rm ref}(v)^w}.
  \label{eq:tilg-q}
\end{equation}
For a pair of tokens, this readout adds $w$ times the change in historical-to-current log odds to the current log odds. We query the readout at the base proposal and use its positive probability gain to score the position:
\begin{equation}
  b_i=\left[q_t(i,\hat x_i)-p_t^{\rm ref}(i,\hat x_i)\right]_+,
  \qquad s_i=c_i+\lambda b_i.
  \label{eq:supp-prob}
\end{equation}
The full vocabulary normalization matters, so a positive candidate log-ratio alone does not determine $b_i$. The parameters $w$ and $\lambda$ control the temporal contrast and its contribution to ranking. Appendix~\ref{app:theory} gives the equivalent probability-ratio form and its first-order interpretation.

\subsection{Persistence and a capped budget select positions}
\label{sec:promotion}
An abrupt support increase can occur before the proposed identity has persisted. The History Gate adds a discrete persistence condition to the continuous score. Let $h_i$ count consecutive steps with the same proposal. For base and extra paths, define
\begin{equation}
  H_i^{j}=\mathbf 1[h_i\ge m_j]\lor\mathbf 1[c_i\ge\tau_{\rm esc}],
  \qquad j\in\{\mathrm{base},\mathrm{extra}\}.
  \label{eq:history-gate-support}
\end{equation}
The confidence escape permits saturated predictions to pass immediately. A confidence floor defines the extra pool, and a per-step cap limits extra promotions. In the gate-before-ranking Dream adapter, the backend's base accept set $\mathcal R_t^{\rm base}$ gives
\begin{align}
  \mathcal B_t&=\{i\in\mathcal R_t^{\rm base}:H_i^{\rm base}=1\},\\
  \mathcal N_t&=\{i\in\mathcal M_t\setminus\mathcal B_t:c_i\ge\tau_{\rm floor},\ H_i^{\rm extra}=1\}.
\end{align}
TACG promotes the highest-scoring candidates under the effective extra budget $k_t$:
\begin{equation}
  \mathcal E_t=\operatorname{TopK}_{i\in\mathcal N_t}(s_i,k_t),\qquad 0\le k_t\le K_{\max}.
\end{equation}
LLaDA preserves its backend ordering: confidence and KL eligibility define the base and extra pools, readiness ranking selects extras under a ratio-capped quota, and a shared History Gate filters their union. The quota uses the raw base-set size and the configured empty-base policy. Both adapters use a progress fallback and write the current base proposal at every accepted position. The cap bounds extra promotions, while the resulting trajectory determines quality and model-call cost. Appendix~\ref{app:pseudocode} gives both adapters explicitly, and Appendix~\ref{app:configurations} lists the experimental settings.

\subsection{The gate changes context timing while writes keep base identity}
TACG computes its gate from logits already available at the current call, so gate computation adds no model forward pass. Its EMA and vocabulary operations scale as $O(BLV)$ for batch size $B$, active length $L$, and vocabulary size $V$. The write rule is local: every accepted position receives its contemporaneous base proposal. Earlier writes can change later contexts, so later proposals and final text can differ across commitment schedules. We therefore evaluate both model-call counts and measured decode latency, and we report logit-reference interventions as trajectory evidence with the local write rule stated explicitly. Appendix~\ref{app:pseudocode} gives the execution order for both adapters, including eligibility pools, quotas, fallback, state updates, and NFE counting; Tables~\ref{tab:registered_vectors} and~\ref{tab:dream_vectors} give the default parameters.

\section{Experiments}
\label{sec:experiments}
\subsection{Evaluation protocol anchors every quality--cost pair}
We evaluate LLaDA-8B and Dream-7B on the complete HumanEval, MBPP, GSM8K,
and MATH500 splits of 164, 257, 1,319, and 500 examples, respectively
~\citep{chen2021evaluatinglargelanguagemodels,
austin2021programsynthesislargelanguage,
cobbe2021trainingverifierssolvemath,
hendrycks2021measuringmathematicalproblemsolving}.
The main comparisons use one completion per example, seed 516, and generation
length 256. We pair each method with its same-model baseline under the same
task and generation budget. LLaDA uses temperature 0; Dream uses temperature
0.2 and top-$p$ 0.95.

We report task accuracy (ACC) and mean main-model forward evaluations (NFE)
per example. Lower NFE indicates fewer model calls. Each accuracy--NFE pair
comes from one complete-task run, and aggregate values average the four
benchmarks equally. LLaDA2-Mini uses separate protocols at generation lengths
1024 and 2048 with MBPP denominator 427. Full protocol details and adapted
baseline schedules appear in Appendix~\ref{app:protocol}. The records in
\texttt{analysis/revision\_table\_data.json},
\texttt{analysis/supplementary\_data/candidate\_results.csv}, and the audited
latency/intervention tables retain task denominators, configuration vectors, and
source-path checks.

Native Confidence commits positions with the backend's current-confidence rule and serves as the same-model baseline. The headline TACG rows use benchmark-specific operating points. We report shared vectors and independently timed vectors as separate settings.

\subsection{Main quality--cost comparison}
\begin{table}[t]
  \centering
  \caption{Accuracy and model-call cost on LLaDA-8B. ACC is accuracy (\%); NFE is mean model calls per example. Mean averages the four tasks equally; bold marks column best. Shaded rows mark TACG; unshaded rows are baselines or controls.}
  \label{tab:revision_main_accuracy}
  \begingroup
  \fontsize{8}{10}\selectfont
  \setlength{\tabcolsep}{2pt}
  \renewcommand{\arraystretch}{1.12}
  \providecolor{oursbg}{RGB}{250,240,232}
  \providecolor{gaincolor}{RGB}{38,100,84}
  \providecommand{\revdelta}[1]{\textcolor{gaincolor}{#1}}
  \begin{tabular}{@{}lrrrrrrrrrr@{}}
    \toprule
    Method & \multicolumn{2}{c}{HumanEval} & \multicolumn{2}{c}{MBPP} & \multicolumn{2}{c}{GSM8K} & \multicolumn{2}{c}{MATH500} & \multicolumn{2}{c}{Mean} \\
    \cmidrule(lr){2-3}\cmidrule(lr){4-5}\cmidrule(lr){6-7}\cmidrule(lr){8-9}\cmidrule(lr){10-11}
    & ACC $\uparrow$ & NFE $\downarrow$ & ACC $\uparrow$ & NFE $\downarrow$ & ACC $\uparrow$ & NFE $\downarrow$ & ACC $\uparrow$ & NFE $\downarrow$ & ACC $\uparrow$ & NFE $\downarrow$ \\
    \midrule
    Native Confidence & 37.20 & 52.28 & 47.08 & 83.34 & 77.10 & 74.26 & 30.60 & 96.48 & 48.00 & 76.59 \\
    AHD & 39.63 & 51.89 & 45.91 & 85.46 & 75.28 & 73.97 & 30.40 & 95.28 & 47.81 & 76.65 \\
    DSB & 39.02 & 50.46 & 48.64 & 81.49 & 76.35 & 72.98 & 33.40 & 95.68 & 49.35 & 75.15 \\
    KLASS & 39.63 & 92.13 & 48.25 & 134.82 & \textbf{77.63} & 134.45 & 30.60 & 152.42 & 49.03 & 128.45 \\
    \rowcolor{oursbg} TACG & \textbf{42.07} & \textbf{45.71} & \textbf{50.97} & \textbf{74.42} & 77.03 & \textbf{67.12} & \textbf{34.00} & \textbf{84.04} & \textbf{51.02} & \textbf{67.82} \\
    \bottomrule
  \end{tabular}
  \endgroup
\end{table}

Table~\ref{tab:revision_main_accuracy} gives the headline LLaDA comparison.
The displayed task-specific TACG points are measured on completed full-task
runs from the candidate roster under the stated single-seed protocol. They
improve accuracy on HumanEval, MBPP, and MATH500 and reduce model calls on all
four benchmarks. Mean accuracy rises
from 48.00\% to 51.02\%, and mean NFE falls from 76.59 to 67.82. Among the
displayed rows, TACG has the highest mean accuracy.
The fixed KLASS row provides an additional GSM8K operating point. The component
and replay studies below test whether the quality--cost change follows from
which positions the decoder commits.

\subsection{Selection and writing contribute differently in a two-by-two intervention}
\label{sec:mechanism}
\begin{table}[!htbp]
  \centering
  \caption{Full task results for the component diagnostic. Mean averages the four complete tasks; Selection is history-guided position choice and Writing changes token values.}
  \label{tab:revision_component_details}
  \begingroup
  \fontsize{8}{10}\selectfont
  \setlength{\tabcolsep}{3.5pt}
  \renewcommand{\arraystretch}{1.12}
  \providecolor{gaincolor}{RGB}{38,100,84}
  \providecommand{\revdelta}[1]{\textcolor{gaincolor}{#1}}
  \begin{tabular}{@{}llrrrrr@{}}
    \toprule
    Method & Metric & HumanEval & MBPP & GSM8K & MATH500 & Mean \\
    \midrule
    Confidence & ACC $\uparrow$ & 37.20 & 47.08 & 77.10 & 30.60 & 48.00 \\
    & NFE $\downarrow$ & 52.28 & 83.34 & 74.26 & 96.48 & 76.59 \\
    \addlinespace[2pt]
    Selection & ACC $\uparrow$ & 39.02 & 50.19 & 76.04 & 32.60 & 49.47\, (\revdelta{+1.47}) \\
    & NFE $\downarrow$ & 48.02 & 74.52 & 63.83 & 85.09 & 67.87\, (\revdelta{-11.39\%}) \\
    \addlinespace[2pt]
    Writing & ACC $\uparrow$ & 37.80 & 49.03 & 75.36 & 31.80 & 48.50\, (\revdelta{+0.50}) \\
    & NFE $\downarrow$ & 50.65 & 82.93 & 74.74 & 97.13 & 76.36\, (\revdelta{-0.30\%}) \\
    \addlinespace[2pt]
    Selection + writing & ACC $\uparrow$ & 40.24 & 49.81 & 75.51 & 32.60 & 49.54\, (\revdelta{+1.55}) \\
    & NFE $\downarrow$ & 46.16 & 74.00 & 64.16 & 85.80 & 67.53\, (\revdelta{-11.83\%}) \\
    \bottomrule
  \end{tabular}
  \endgroup
\end{table}

The component study in Table~\ref{tab:revision_component_details} uses an
independent across-task LLaDA batch, so its absolute values differ from the
headline task-specific rows. The history-guided selection policy is composite:
it combines temporal ranking, capped extra promotion, persistence and escape
checks, and progress fallback. Relative to its matched control, the selection arm changes mean accuracy by +1.47 points and mean NFE by -11.39\%; the guided-token-writing arm
changes accuracy by +0.50 points and NFE by -0.30\%. The combined
variant reaches 49.54\% accuracy with 67.53 mean NFE. The diagnostic shows a
larger observed shift in the composite selection arm under this intervention.
The factorization supports placing temporal evidence in the selector while the
base decoder supplies the written token.

\begin{table*}[t]
  \centering

  \begin{minipage}[t]{0.30\textwidth}
    \vspace{0pt}
    \centering
    \captionof{table}{Matched per-event commit-count selector replay. Both selectors use identical per-sample NFE; selected positions differ.}
    \label{tab:revision_equal_budget}

    \begingroup
    \fontsize{9}{11}\selectfont
    \renewcommand{\arraystretch}{1.12}
    \resizebox{\linewidth}{!}{%
    \begin{tabular}{@{}lrr@{}}
      \toprule
      Selector & ACC $\uparrow$ & NFE $\downarrow$ \\
      \midrule
      History-guided & 49.42 & 62.79 \\
      Confidence-ranked & 41.16 & 62.79 \\
      \bottomrule
    \end{tabular}%
    }

    \endgroup
  \end{minipage}
  \hfill
  \begin{minipage}[t]{0.68\textwidth}
    \vspace{0pt}
    \centering
    \captionof{table}{Per-task results under matched per-event commit counts.}
    \label{tab:revision_equal_budget_tasklevel}

    \begingroup
    \fontsize{8.5}{10}\selectfont
    \renewcommand{\arraystretch}{1.08}
    \resizebox{\linewidth}{!}{%
    \begin{tabular}{@{}llrrrrr@{}}
      \toprule
      Selector & Metric & HumanEval & MBPP & GSM8K & MATH500 & Mean \\
      \midrule
      History-guided
      & ACC $\uparrow$ & 39.02 & 50.19 & 76.27 & 32.20 & 49.42 \\
      & NFE $\downarrow$ & 44.02 & 69.39 & 58.33 & 79.43 & 62.79 \\
      \addlinespace[2pt]
      Confidence-ranked
      & ACC $\uparrow$ & 34.76 & 38.52 & 67.55 & 23.80 & 41.16 \\
      & NFE $\downarrow$ & 44.02 & 69.39 & 58.33 & 79.43 & 62.79 \\
      \bottomrule
    \end{tabular}%
    }

    \endgroup
  \end{minipage}
  \vspace{-1.5em}
\end{table*}

\subsection{History selection matters under matched per-event commit counts}

The replay tests position choice under matched per-event commit counts. The
confidence-ranked control reproduces the reference's total commit count at every
sample, block, and event, so both selectors use exactly the same per-sample NFE.
History-guided selection reaches 49.42\% accuracy versus 41.16\% for
confidence ranking, a difference of 8.27 percentage points across the four tasks.

\begin{figure}[htbp]
    \centering
    \vspace{-1em}
    \includegraphics[width=1\linewidth]
    {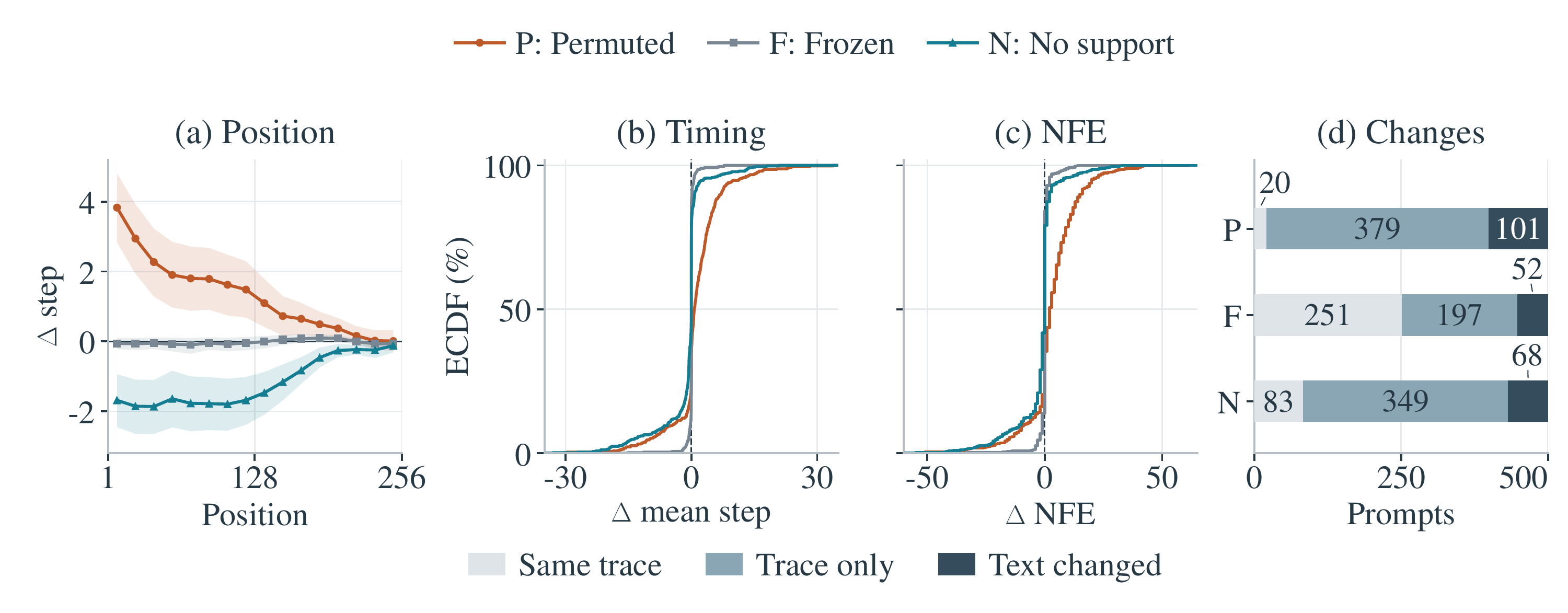}
    \vspace{-2.8em}
    \caption{
    Dream MATH500 intervention diagnostic on 500 paired prompts. (a) Mean commit-step changes across 16-position bins with pointwise 95\% intervals; (b,c) ECDFs of prompt-level mean commit-step and NFE changes; (d) joint changes in position-wise $(\text{step},\text{source})$ traces and final text.
    }
    \label{fig:history_distributions}
\end{figure}

\subsection{Logit-reference interventions change trajectories while local writes stay anchored}

The four arms use the observed EMA reference, a frozen reference, no temporal
support, and a within-sample position-permuted reference. The no-support arm
removes the temporal-support branch; the other intervention arms alter its
reference. Confidence and the proposal-streak gate remain fixed. Relative to the
observed reference, the three interventions change 417, 249, and 480 of 500
commit traces, and 68, 52, and 101 final texts, respectively. All 512,000
recorded writes match their local base proposal. All three controls have paired
accuracy intervals; together, these controls provide a
trajectory-sensitivity view of the commit path and final text.

\subsection{The measured roster covers quality--NFE operating points}
\begin{figure}[htbp]
  \centering
  \includegraphics[width=0.8\linewidth]{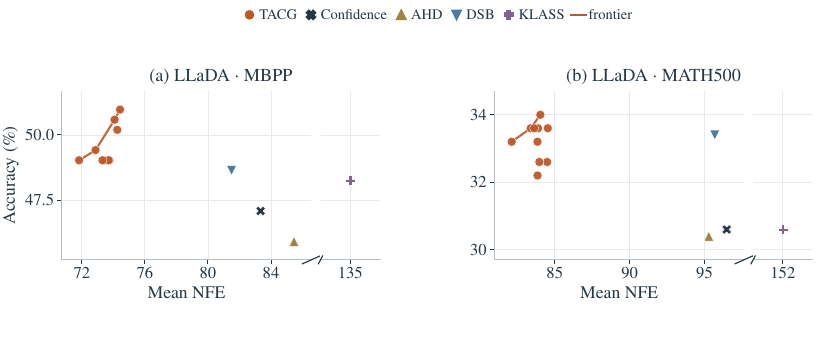}
  \vspace{-1em}
  \caption{
  \textbf{Measured quality--NFE operating points on LLaDA.}
  We compare TACG settings with Native Confidence, AHD, DSB, and KLASS.
  The line connects the measured non-dominated TACG points.
  The appendix reports the controlled $\lambda$ slice and Dream sweep.
  }

  \label{fig:quality_nfe_selected}
\end{figure}
Figure~\ref{fig:quality_nfe_selected} shows the measured LLaDA quality--NFE
roster. Each orange point is a complete-task evaluation and the line joins the
measured non-dominated points. The main table selects one operating point per
benchmark; the appendix reports the complete LLaDA and Dream candidate rosters.

The sensitivity panels vary one control at a time on LLaDA HumanEval. The
support-weight slice covers $\lambda=0.15$--$0.75$ at $w=0.35$; the headline
HumanEval operating point uses $\lambda=2.00$ and is outside this slice. The
prompt-level panel evaluates six History Gate settings on the same 164 examples.
\begin{figure}[htbp]
  \centering
  \begin{minipage}[t]{.48\linewidth}
    \centering
    \includegraphics[width=0.9\linewidth]{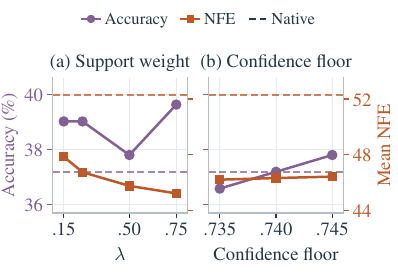}
    \caption{\textbf{Single-parameter sensitivity on LLaDA HumanEval}. We vary $\lambda$ from 0.15 to 0.75 at $w=0.35$ and the confidence floor from 0.735 to 0.745, one parameter at a time. Dashed lines denote Native Confidence; these slices are separate from the main HumanEval operating point.}
    \label{fig:lambda_floor}
  \end{minipage}\hfill
  \begin{minipage}[t]{.48\linewidth}
    \centering
    \includegraphics[width=0.9\linewidth]{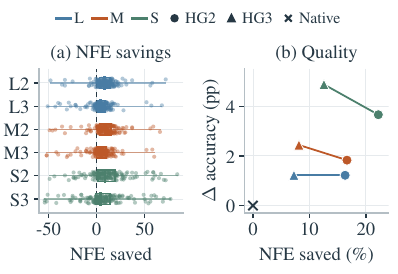}
    \caption{
    Prompt-level call changes on HumanEval across six History Gate settings (984 paired records). L/M/S label the low, moderate, and strong support vectors in Table~\ref{tab:hg_window}; 2/3 denote the persistence windows. (a) NFE saved is Native minus TACG.
    (b) Mean relative NFE saving and accuracy change.
    }
    \label{fig:prompt_efficiency}
  \end{minipage}
\end{figure}

\subsection{Adapter-specific tests span backbones and lengths}
\begin{table}[t]
  \centering
  \caption{Accuracy and model-call cost across backbones. Mean averages the four tasks equally; Dream uses MBPP257 and LLaDA2-Mini uses MBPP427. Shaded rows mark TACG.}
  \label{tab:revision_shared_transfer}
  \begingroup
  \fontsize{8}{10}\selectfont
  \setlength{\tabcolsep}{2pt}
  \renewcommand{\arraystretch}{1.12}
  \providecolor{oursbg}{RGB}{250,240,232}
  \providecolor{gaincolor}{RGB}{38,100,84}
  \providecommand{\revdelta}[1]{\textcolor{gaincolor}{#1}}
  \begin{tabular}{@{}lrrrrrrrrrr@{}}
    \toprule
    Method & \multicolumn{2}{c}{HumanEval} & \multicolumn{2}{c}{MBPP} & \multicolumn{2}{c}{GSM8K} & \multicolumn{2}{c}{MATH500} & \multicolumn{2}{c}{Mean} \\
    \cmidrule(lr){2-3}\cmidrule(lr){4-5}\cmidrule(lr){6-7}\cmidrule(lr){8-9}\cmidrule(lr){10-11}
    & ACC $\uparrow$ & NFE $\downarrow$ & ACC $\uparrow$ & NFE $\downarrow$ & ACC $\uparrow$ & NFE $\downarrow$ & ACC $\uparrow$ & NFE $\downarrow$ & ACC $\uparrow$ & NFE $\downarrow$ \\
    \midrule
    \multicolumn{11}{@{}l}{\textit{LLaDA-8B}} \\
    Native Confidence & 37.20 & 52.28 & 47.08 & 83.34 & \textbf{77.10} & 74.26 & 30.60 & 96.48 & 48.00 & 76.59 \\
    \rowcolor{oursbg} TACG & \textbf{42.07} & \textbf{45.71} & \textbf{50.97} & \textbf{74.42} & 77.03 & \textbf{67.12} & \textbf{34.00} & \textbf{84.04} & \textbf{51.02} & \textbf{67.82} \\
    \addlinespace[3pt]
    \multicolumn{11}{@{}l}{\textit{Dream-7B}} \\
    Native Confidence & 47.56 & 51.19 & \textbf{63.42} & \textbf{82.46} & 73.16 & 74.63 & 39.00 & 94.60 & 55.79 & \textbf{75.72} \\
    \rowcolor{oursbg} TACG & \textbf{50.00} & \textbf{50.80} & 63.04 & 93.53 & \textbf{73.62} & \textbf{74.26} & \textbf{39.60} & \textbf{91.14} & \textbf{56.56} & 77.43 \\
    \addlinespace[3pt]
    \multicolumn{11}{@{}l}{\textit{LLaDA2-Mini}} \\
    Baseline & 84.15 & \textbf{67.38} & 80.09 & 77.93 & 87.64 & \textbf{147.90} & 64.80 & 294.77 & 79.17 & 146.99 \\
    \rowcolor{oursbg} TACG & \textbf{85.98} & 69.66 & \textbf{80.56} & \textbf{75.36} & \textbf{88.17} & 148.11 & \textbf{67.40} & \textbf{290.79} & \textbf{80.53} & \textbf{145.98} \\
    \bottomrule
  \end{tabular}
  \vspace{-1em}
  \endgroup
\end{table}

Table~\ref{tab:revision_shared_transfer} applies the same temporal-support framework, adapted to three decoder families. On the tested Dream task-specific operating points, TACG improves both accuracy and call count on HumanEval, GSM8K, and MATH500; MBPP supplies the complementary operating point under the same official adapter. On LLaDA2-Mini, accuracy improves on all four tasks at both 1024 and 2048 generated positions, with mean NFE decreasing from 146.99 to 145.98 and from 195.86 to 195.19. The table below lists the matched task-level results; the two generation budgets remain separate because they correspond to different block trajectories.
\begin{table}[t]
  \centering
  \vspace{-1em}
  \caption{LLaDA2-Mini at two generation budgets. Each panel matches the generation budget; MBPP uses 427 examples and shaded rows mark TACG.}
  \label{tab:mini_length_comparison}
  \begingroup
  \fontsize{8}{10}\selectfont
  \setlength{\tabcolsep}{2pt}
  \renewcommand{\arraystretch}{1.12}
  \providecolor{oursbg}{RGB}{250,240,232}
  \providecolor{gaincolor}{RGB}{38,100,84}
  \providecommand{\revdelta}[1]{\textcolor{gaincolor}{#1}}
  \begin{tabular}{@{}lrrrrrrrrrr@{}}
    \toprule
    Method & \multicolumn{2}{c}{HumanEval} & \multicolumn{2}{c}{MBPP} & \multicolumn{2}{c}{GSM8K} & \multicolumn{2}{c}{MATH500} & \multicolumn{2}{c}{Mean} \\
    \cmidrule(lr){2-3}\cmidrule(lr){4-5}\cmidrule(lr){6-7}\cmidrule(lr){8-9}\cmidrule(lr){10-11}
    & ACC $\uparrow$ & NFE $\downarrow$ & ACC $\uparrow$ & NFE $\downarrow$ & ACC $\uparrow$ & NFE $\downarrow$ & ACC $\uparrow$ & NFE $\downarrow$ & ACC $\uparrow$ & NFE $\downarrow$ \\
    \midrule
    \multicolumn{11}{@{}l}{\textit{Generation length 1024}} \\
    Baseline & 84.15 & \textbf{67.38} & 80.09 & 77.93 & 87.64 & \textbf{147.90} & 64.80 & 294.77 & 79.17 & 146.99 \\
    \rowcolor{oursbg} TACG & \textbf{85.98} & 69.66 & \textbf{80.56} & \textbf{75.36} & \textbf{88.17} & 148.11 & \textbf{67.40} & \textbf{290.79} & \textbf{80.53} & \textbf{145.98} \\
    \addlinespace[3pt]
    \multicolumn{11}{@{}l}{\textit{Generation length 2048}} \\
    Baseline & 84.15 & \textbf{71.51} & 80.09 & \textbf{78.12} & 88.32 & 176.97 & 75.20 & 456.83 & 81.94 & 195.86 \\
    \rowcolor{oursbg} TACG & \textbf{85.37} & 78.32 & \textbf{80.33} & 79.72 & \textbf{88.63} & \textbf{171.18} & \textbf{76.40} & \textbf{451.52} & \textbf{82.68} & \textbf{195.19} \\
    \bottomrule
  \end{tabular}
  \vspace{-0.5em}
  \endgroup
\end{table}

\subsection{Measured decode cost is separate from NFE}
The default rows use benchmark-specific points; the appendix reports one shared
vector and a separate timing-study vector. We measure request-level decoding
behavior instead of inferring wall-clock gains from NFE. The paired History Gate
study changes only the persistence window, and the latency table uses matched
requests under its own configuration.
\begin{table}[t]
  \centering
  \begin{minipage}[t]{0.50\linewidth}
    \vspace{0pt}
    \centering\fontsize{8}{10}\selectfont
    \captionof{table}{History Gate window on HumanEval. Each pair changes only the minimum proposal streak; the three rows use distinct support settings.}
    \label{tab:hg_window}
    \setlength{\tabcolsep}{2.7pt}\renewcommand{\arraystretch}{1.16}
    \begin{tabular}{@{}crrrr@{}}
      \toprule
      & \multicolumn{2}{c}{HG = 2} & \multicolumn{2}{c}{HG = 3} \\
      \cmidrule(lr){2-3}\cmidrule(lr){4-5}
      $(w,\lambda)$ & ACC & NFE & ACC & NFE \\
      \midrule
      (0.30, 0.05) & \textbf{38.41} & \textbf{43.75} & \textbf{38.41} & 48.49 \\
      (0.35, 0.10) & 39.02 & \textbf{43.58} & \textbf{39.63} & 48.02 \\
      (0.38, 2.00) & 40.85 & \textbf{40.68} & \textbf{42.07} & 45.71 \\
      \bottomrule
    \end{tabular}
  \end{minipage}\hfill
  \begin{minipage}[t]{0.47\linewidth}
    \vspace{0pt}
    \centering\fontsize{9}{11}\selectfont
    \captionof{table}{Matched request cost at two HG windows on H200. Each setting uses the same 80 timing requests.}
    \label{tab:revision_c0_l12_systems}
    \setlength{\tabcolsep}{4pt}\renewcommand{\arraystretch}{1.16}
    \begin{tabular}{@{}lrr@{}}
      \toprule
      Metric & HG = 2 & HG = 3 \\
      \midrule
      NFE & \textbf{85.05} & 90.10 \\
      Decode (ms) & \textbf{2145.88} & 2279.48 \\
      Request (ms) & \textbf{2334.90} & 2427.97 \\
      Wall speedup & \textbf{1.040$\times$} & 1.000$\times$ \\
      \bottomrule
    \end{tabular}
  \end{minipage}
\end{table}

\begin{table}[!htbp]
  \centering
  \vspace{-1em}
  \caption{Matched decoding latency for alternate LLaDA parameter settings. Mean synchronized decode time (ms) on H200; speedup is the four-task mean ratio.}
  \label{tab:revision_main_latency}
  \begingroup
  \fontsize{9}{11}\selectfont
  \setlength{\tabcolsep}{2pt}
  \renewcommand{\arraystretch}{1.12}
  \providecolor{gaincolor}{RGB}{38,100,84}
  \providecommand{\revdelta}[1]{\textcolor{gaincolor}{#1}}
  \begin{tabular}{@{}llrrrrrr@{}}
    \toprule
    Method & Setting & HumanEval & MBPP & GSM8K & MATH500 & Mean ms $\downarrow$ & Speedup $\uparrow$ \\
    \midrule
    Native Confidence & Reference & 1391.3 & 2073.7 & 1975.5 & 2341.0 & 1945.4 & 1.000$\times$ \\
    TACG & Across-task & 1330.5 & 1949.8 & 1784.9 & 2119.7 & 1796.2 & 1.083$\times$ \\
    TACG & Timing study & 1383.0 & 1780.1 & 1836.4 & 2119.7 & 1779.8 & 1.093$\times$ \\
    \bottomrule
  \end{tabular}
  \endgroup
\end{table}

The fully profiled timing-study vector gives a 1.093$\times$ decode-speed ratio
on LLaDA, with mean synchronized decode times of 1779.8 ms and 1945.4 ms. The
timing-study vector is a separate fully profiled quality--NFE point with 50.47\%
quality and 66.82 mean NFE. Appendix~\ref{app:length_batch_scaling} reports throughput and memory
on the tested length--batch grid.

\section{Conclusion}
\label{sec:conclusion}
TACG uses the denoising trajectory to decide when a base proposal should enter the context. Its temporal-support and persistence signals change position selection while retaining the current proposal at every write. Complete-task comparisons on the tested single-seed LLaDA roster show accuracy gains on three of four benchmarks with fewer calls on all four, and controlled interventions show observed selector differences under matched per-event commit counts and identical per-sample NFE. Dream, LLaDA2-Mini, and the system profile characterize tested operating points across backbones, lengths, and throughput configurations.

\clearpage

\bibliographystyle{iclr2027_conference}
\bibliography{refs,analysis/baseline_references}
\clearpage
\appendix

\clearpage
\section{Decoding algorithm}
\label{app:pseudocode}
The listings give the shared TACG computations and the LLaDA and Dream selection rules used in our experiments. Each listing targets one sequence; we apply selection independently to each active batch row. \texttt{topk(score, candidates, k)} returns a Boolean mask containing the highest-scoring $\min(k,|\mathrm{candidates}|)$ eligible positions, and returns an empty mask when $k=0$.

\paragraph{Shared temporal state and readout.}
The state contains the EMA logits, the previous base proposal, and its consecutive-agreement count. The first observation gives zero temporal support. The historical reference is read before the current logits enter the EMA.
\begin{lstlisting}[style=tacgpseudocode]
class TACGState:
    def __init__(self):
        self.ref = None
        self.previous = None
        self.streak = None


def tacg_readout(logits, history_mask, state, cfg):
    p = softmax(logits, dim=-1)
    proposal = argmax(p, dim=-1)
    confidence = gather(p, proposal)
    if state.previous is None:
        streak = ones_like(proposal)
    else:
        same = proposal == state.previous
        streak = where(same, state.streak + 1, 1)
    streak = where(history_mask, streak, 0)
    ref = logits if state.ref is None else state.ref
    q = softmax(logits + cfg.w * (logits - ref), dim=-1)
    p_ref = softmax(ref, dim=-1)
    support = clamp(gather(q, proposal)
                    - gather(p_ref, proposal), min=0)
    if state.ref is None:
        support = zeros_like(confidence)
    score = confidence + cfg.support_lambda * support
    return p, proposal, confidence, streak, score, ref


def update_state(state, logits, proposal, streak, ref, beta):
    state.ref = detach(beta * ref + (1 - beta) * logits)
    state.previous = detach(proposal)
    state.streak = detach(streak)
\end{lstlisting}

\clearpage
\paragraph{Base eligibility and backend selection.}
Let \texttt{kl\_ok} be true for a confidence-only gate. For an active KLASS gate it requires the backend's configured window of consecutive values $D_{\rm KL}(p_t\|p_{t-1})$ to fall below its KL threshold, using the same zero-initialized posterior/window and numerical offset as the backend. LLaDA retains this eligibility check for both raw base candidates and extra candidates. Its ratio cap uses the number of raw base candidates, and its History Gate filters the combined candidate set after extra selection. Dream applies separate persistence thresholds to the base and extra paths and uses a fixed extra cap.

\begin{lstlisting}[style=tacgpseudocode]
def llada_select(mask, confidence, streak, score, kl_ok,
                 scheduled_k, final_step, cfg):
    raw_base = mask & kl_ok & (confidence > cfg.tau_conf)
    eligible = (mask & kl_ok
                & (confidence > cfg.tau_floor)
                & (confidence <= cfg.tau_conf))
    k = min(cfg.K_max, ceil(cfg.extra_ratio
                           * max(count(raw_base), 1)))
    if not any(raw_base) and not cfg.allow_empty_base:
        k = 0
    extra = topk(score, eligible, k)
    candidates = raw_base | extra
    if not any(candidates):
        candidates = topk(score, mask, scheduled_k)
    history_ok = ((streak >= cfg.m_base)
                  | (confidence >= cfg.tau_escape))
    commit = candidates & history_ok & mask
    if not any(commit) and any(candidates):
        commit = topk(score, candidates, 1)
    elif not any(commit) and final_step:
        commit = topk(score, mask, scheduled_k)
    return commit


def dream_select(logits, mask, confidence, streak, score,
                 kl_ok, cfg):
    raw_base = mask & kl_ok & (confidence >= cfg.tau_conf)
    escape = confidence >= cfg.tau_escape
    base = raw_base & ((streak >= cfg.m_base) | escape)
    eligible = (mask & ~base
                & (confidence >= cfg.tau_floor)
                & ((streak >= cfg.m_extra) | escape))
    extra = topk(score, eligible, cfg.K_max)
    commit = base | extra
    if not any(commit) and any(mask):
        # Same position selection as the Dream baseline.
        fallback_conf, _ = sample_tokens(
            logits[mask], temperature=cfg.temperature,
            top_p=cfg.top_p, top_k=cfg.top_k)
        fallback_score = full_like(confidence, -inf)
        fallback_score[mask] = fallback_conf
        commit = topk(fallback_score, mask, 1)
    return commit
\end{lstlisting}
The Dream fallback samples from the backend's temperature-scaled and truncated posterior to score positions; we discard the sampled token. The write below always uses the current base-posterior argmax. In LLaDA, the same persistence threshold applies to base and extra candidates, and HG rejects an extra candidate without replacing it with a lower-ranked candidate. We account for mandatory progress separately from the extra cap.

\clearpage
\paragraph{Block loop, writes, and model-call counting.}
LLaDA uses its configured blocks, retains temporal state across blocks, and tracks proposal streaks over the full sequence. Dream treats the generation region as one active block and updates streaks at masked positions. For a block with $n$ initially masked positions and $S$ scheduled steps, LLaDA assigns $\lfloor n/S\rfloor+\mathbf 1[t<n\bmod S]$ fallback positions to step $t$. The following loop shows where these state updates and the NFE counter occur; the loop computes \texttt{kl\_ok} from the same forward result. The reported confidence settings keep the KL check open; active KLASS configurations retain the window check shown below.
\begin{lstlisting}[style=tacgpseudocode]
x = concatenate(prompt, full(generation_length, MASK))
state, nfe = TACGState(), 0
previous_p, kl_history = None, zeros(len(x), cfg.kl_window)
for block in generation_blocks:
    initial_count = count((x == MASK) & block)
    steps = cfg.steps_for_block(block)
    for t in range(steps):
        mask = (x == MASK) & block
        if not any(mask):
            break
        logits = model(x).logits
        nfe += 1                         # one actual model call
        logits = align_backend_logits(logits)
        history_mask = ones_like(mask) if cfg.llada else mask
        p, proposal, conf, streak, score, ref = tacg_readout(
            logits, history_mask, state, cfg)
        kl_ok = ones_like(mask)
        if cfg.use_kl:
            if previous_p is None:
                previous_p = zeros_like(p)
            d = sum(p * (log(p + 1e-12)
                         - log(previous_p + 1e-12)), dim=-1)
            kl_history = append_and_drop_oldest(kl_history, d)
            kl_ok = all(kl_history < cfg.tau_kl, dim=-1)
            if cfg.llada and t < cfg.kl_window - 1:
                kl_ok = zeros_like(mask)
            previous_p = detach(p)
        if cfg.llada:
            budget = (initial_count // steps
                      + (t < initial_count % steps))
            commit = llada_select(mask, conf, streak, score,
                                  kl_ok, budget, t == steps - 1, cfg)
        else:
            commit = dream_select(logits, mask, conf, streak,
                                  score, kl_ok, cfg)
        x[commit] = proposal[commit]      # base, extra, and fallback
        update_state(state, logits, proposal, streak, ref, cfg.beta)
        mask = mask & ~commit
return x, nfe
\end{lstlisting}
\texttt{align\_backend\_logits} applies the backbone's output-position convention (Dream shifts its logits by one position). TACG's readout and selection need no additional model evaluation. The LLaDA2-Mini implementation retains its block and cache schedule; its reported decoding NFE also includes cache-refresh model calls and excludes prompt prefill. Experimental settings appear in Appendix~\ref{app:configurations}.

\section{Protocols and configurations}
\label{app:protocol}
\paragraph{Complete task evaluation.}
We use generation length 256 for the main LLaDA-8B/Dream-7B tables and ordered, complete task splits with one generated completion per example. We evaluate LLaDA2-Mini separately at lengths 1024 and 2048; baseline and TACG share the generation budget within each comparison. We score code-task accuracy on the saved completions and compute NFE from per-example model-call counters. Accuracy, mean NFE, and latency have distinct denominators: the first two cover complete tasks, whereas matched decode profiles use repeated timing requests. We average accuracy and NFE across the four tasks within each panel.

On GSM8K, the LLaDA Native Confidence and TACG rows use the same 1,319 test questions, checkpoint, prompts, and answer parser, with generation length 256, block length 64, and step budget 256. Their saved per-question results give 1,017 and 1,016 correct answers and mean NFE 74.26 and 67.12, respectively. We aggregate both metrics over the complete test split.

\paragraph{Adapted baselines.}
Native Confidence and fixed KLASS serve as separate gate families; KLASS uses confidence threshold 0.9, KL threshold 0.01, and history length 2 on all tasks. AHD and DSB use the authors' implementations adapted to the evaluation protocol with their method-specific schedules. AHD uses a fixed KL threshold of 0.01 across tasks; DSB uses its greedy scheduler with caching disabled. NFE counts main-model evaluations, whose computational cost can differ across implementations.

\paragraph{Operating-point selection.}
The default TACG row uses benchmark-specific operating points. The across-task ablation uses one TACG vector for all tasks within each model.

\subsection{Hyperparameter settings}
\label{app:configurations}
The following tables list the registered vectors and full candidate rosters. Both backbones use EMA decay $\beta=0.95$. LLaDA uses HG streak 3, at most two extra commits, extra ratio 0.25, and empty-base promotion. For Dream, $m_b$ and $m_e$ denote the base and extra-path streak requirements, and $K$ is the maximum extra-commit budget. The main Dream row uses one benchmark-specific accuracy--NFE operating point per task with the same current TACG adapter; the accompanying machine-readable records retain the corresponding raw paths and receipts.
\begin{table}[htbp]
  \centering\fontsize{9}{11}\selectfont
  \caption{TACG parameters on LLaDA. The main comparison uses Default; the Timing study setting uses Default vectors for HumanEval/GSM8K and the separate vectors below for MBPP/MATH500. Across-task uses one vector for all tasks. All use confidence threshold 0.9, EMA decay 0.95, at most two extra commits, and extra ratio 0.25. The settings use empty-base promotion.}
  \label{tab:registered_vectors}
  \setlength{\tabcolsep}{3.5pt}\renewcommand{\arraystretch}{1.12}
  \begin{tabular}{@{}llrrrrr@{}}
    \toprule
    Setting & Task & $w$ & $\lambda$ & $\tau_{\rm floor}$ & HG & $\tau_{\rm esc}$ \\
    \midrule
    Across-task & All tasks & 0.35 & 0.1 & 0.74 & 3 & 0.95 \\
    Default & HumanEval & 0.38 & 2 & 0.74 & 3 & 0.96 \\
    Default & MBPP & 0.35 & 0.15 & 0.74 & 3 & 0.95 \\
    Default & GSM8K & 0.3 & 0.05 & 0.8 & 3 & 0.95 \\
    Default & MATH500 & 0.365 & 0.35 & 0.745 & 3 & 0.95 \\
    Timing study & MBPP & 0.35 & 0.1 & 0.74 & 2 & 0.95 \\
    Timing study & MATH500 & 0.35 & 0.1 & 0.74 & 3 & 0.95 \\
    \bottomrule
  \end{tabular}
\end{table}

\begin{table}[htbp]
  \centering\fontsize{9}{11}\selectfont
  \caption{TACG parameters on Dream. Default gives the benchmark-specific operating points used in the main comparison; Across-task uses one vector for all tasks. All rows use the same TACG adapter with EMA decay 0.95. Base confidence is 0.95 on MBPP and 0.90 on the other tasks; $m_b$ and $m_e$ give the base and extra-path streaks, and $K$ caps extra commits.}
  \label{tab:dream_vectors}
  \setlength{\tabcolsep}{3pt}\renewcommand{\arraystretch}{1.12}
  \begin{tabular}{@{}llrrrrrrr@{}}
    \toprule
    Setting & Task & $w$ & $\lambda$ & $\tau_{\rm floor}$ & $m_b$ & $m_e$ & $\tau_{\rm esc}$ & $K$ \\
    \midrule
    Across-task & All tasks & 0.3 & 0.02 & 0.885 & 5 & 3 & 0.99 & 1 \\
    Default & HumanEval & 0.3 & 0.02 & 0.895 & 1 & 1 & 1 & 1 \\
    Default & MBPP & 0.3 & 0.02 & 0.89 & 5 & 3 & 0.99 & 2 \\
    Default & GSM8K & 0.225 & 0.015 & 0.899 & 1 & 1 & 1 & 1 \\
    Default & MATH500 & 0.35 & 0.1 & 0.88 & 3 & 1 & 0.95 & 2 \\
    \bottomrule
  \end{tabular}
\end{table}

\begin{table}[htbp]
  \centering
  \begin{minipage}[t]{0.38\linewidth}
    \vspace{0pt}
    \centering\fontsize{9}{11}\selectfont
    \captionof{table}{LLaDA sweep: each row is one parameter vector, applied to all four tasks.}
    \label{tab:sweep_llada}
    \setlength{\tabcolsep}{2.7pt}\renewcommand{\arraystretch}{1.16}
    \begin{tabular}{@{}rrrr@{}}
      \toprule
      $w$ & $\lambda$ & $\tau_{\rm floor}$ & $\tau_{\rm esc}$ \\
      \midrule
      0.35 & 0.15 & 0.74 & 0.95 \\
      0.35 & 0.25 & 0.74 & 0.95 \\
      0.365 & 0.25 & 0.74 & 0.95 \\
      0.365 & 0.35 & 0.74 & 0.95 \\
      0.365 & 0.35 & 0.745 & 0.95 \\
      0.365 & 0.35 & 0.74 & 0.96 \\
      0.38 & 0.35 & 0.74 & 0.95 \\
      0.35 & 0.5 & 0.74 & 0.95 \\
      0.365 & 0.35 & 0.735 & 0.95 \\
      0.35 & 0.75 & 0.74 & 0.95 \\
      \bottomrule
    \end{tabular}
  \end{minipage}\hfill
  \begin{minipage}[t]{0.59\linewidth}
    \vspace{0pt}
    \centering\fontsize{9}{11}\selectfont
    \captionof{table}{Dream sweep: each row is one parameter vector, applied to all four tasks.}
    \label{tab:sweep_dream}
    \setlength{\tabcolsep}{2.7pt}\renewcommand{\arraystretch}{1.16}
    \begin{tabular}{@{}rrrrrrr@{}}
      \toprule
      $w$ & $\lambda$ & $\tau_{\rm floor}$ & $m_b$ & $m_e$ & $\tau_{\rm esc}$ & $K$ \\
      \midrule
      0.3 & 0.02 & 0.885 & 5 & 3 & 0.99 & 1 \\
      0.3 & 0.02 & 0.885 & 4 & 3 & 0.99 & 1 \\
      0.3 & 0.02 & 0.88 & 4 & 3 & 0.99 & 1 \\
      0.3 & 0.02 & 0.88 & 5 & 2 & 0.99 & 1 \\
      0.3 & 0.02 & 0.875 & 5 & 3 & 0.99 & 1 \\
      0.3 & 0.02 & 0.88 & 5 & 3 & 0.985 & 1 \\
      0.3 & 0.035 & 0.88 & 5 & 3 & 0.99 & 1 \\
      0.275 & 0.02 & 0.88 & 5 & 3 & 0.99 & 1 \\
      0.225 & 0.015 & 0.89 & 5 & 3 & 0.99 & 1 \\
      0.3 & 0.02 & 0.89 & 5 & 3 & 0.99 & 2 \\
      \bottomrule
    \end{tabular}
  \end{minipage}
\end{table}

Tables~\ref{tab:sweep_llada} and~\ref{tab:sweep_dream} give the complete candidate vectors used by the operating-point analysis.

\section{Complete candidate results}
\label{app:more-experiments}
\subsection{LLaDA2-Mini at longer generation lengths}
\label{app:mini}
Table~\ref{tab:mini_length_comparison} reports both Mini generation budgets, including the 1024-token setting used in Table~\ref{tab:revision_shared_transfer}. The comparisons use complete HumanEval, MBPP, GSM8K, and MATH500 splits with 164, 427, 1,319, and 500 examples. Mini uses the historical MBPP427 protocol, separate from the MBPP257 split used for LLaDA-8B and Dream; NFE counts decoding calls across blocks and cache refreshes, excluding prompt prefill. We evaluate both lengths with task-specific TACG settings, with the 2048-token History Gate enabled for HumanEval and MBPP. The two generation budgets are separate operating points.

\subsection{Complete quality--NFE candidate coverage}
\label{app:operating_points}
\begin{figure}[htbp]
  \centering
  \includegraphics[width=\linewidth]{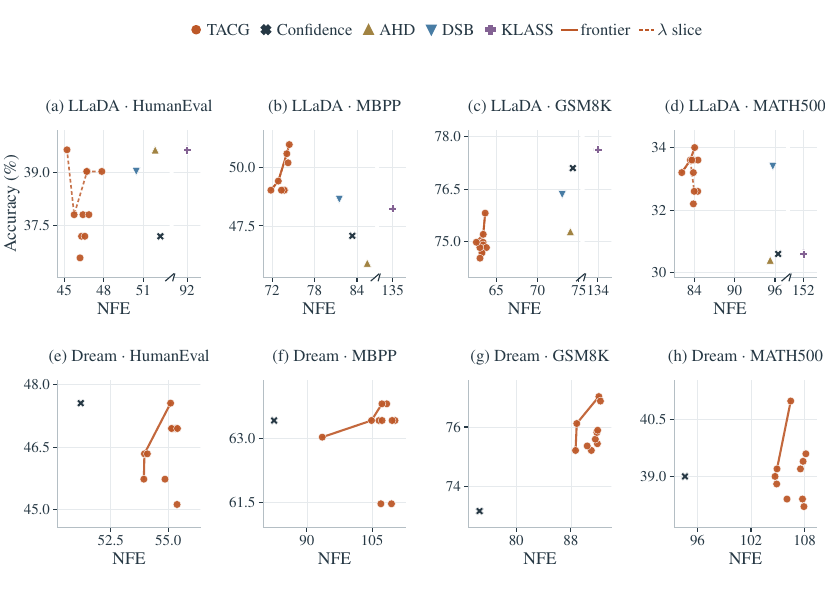}
  \caption{Complete two-backbone candidate sweep: 80 TACG task results across four benchmarks and the available baseline overlays. Each panel retains all ten shared-vector candidates; the LLaDA panels include Native Confidence, AHD, DSB, and the fixed-vector KLASS control, while Dream includes its audited Native Confidence overlay. The solid orange line joins each panel's non-dominated candidates, as in Figure~\ref{fig:quality_nfe_selected}. The dashed orange segments in the LLaDA panels connect the controlled four-point $\lambda$ slice, $\lambda \in \{0.15, 0.25, 0.50, 0.75\}$ with all remaining settings fixed, in parameter order; the two lines coincide on MBPP and differ elsewhere. Table~\ref{tab:revision_shared_transfer} reports the current Dream task-specific rows; the main-text LLaDA panels show selected views of this complete sweep.}
  \label{fig:quality_nfe_complete}
\end{figure}

We evaluate the ten declared configurations per backbone on all four complete tasks, yielding 80 candidate task results and 44,800 generated samples. Each configuration shares its TACG vector across tasks within its model, with the audited task-specific base-gate presets retained. Figure~\ref{fig:quality_nfe_complete} gives the complete two-backbone candidate view; the parameter tables and machine-readable candidate table retain the corresponding vectors and task results. The default rows of Tables~\ref{tab:revision_main_accuracy} and~\ref{tab:revision_shared_transfer} use the benchmark-specific vectors of Tables~\ref{tab:registered_vectors} and~\ref{tab:dream_vectors}. The LLaDA MBPP and MATH500 defaults and the Dream MBPP default coincide with sweep members; the remaining defaults are separate complete-task runs with the same adapters.

\section{Mechanism audits and sensitivity}
\label{app:mechanism_details}
\subsection{Selection and token-writing interventions}
The selection intervention enables the shared history-guided position policy: temporal-bonus ranking, capped extra promotion, proposal-streak/escape checks, and progress fallback. The writing intervention changes the written token to the guided-logit argmax. Their four-way comparison shares inputs and saved configuration fields apart from the algorithm switch. The study uses matched runs for each intervention.

\subsection{Replayed per-event commit counts}
The history-guided reference run supplies 141,707 commit events per arm. The confidence-ranked control reproduces the total committed-token count at every aligned sample/block/event and therefore the exact per-sample NFE. Confidence ranking selects current top-confidence positions; the detailed table retains the same two selectors with their task-level values.
\begin{table}[!htbp]
  \centering
  \caption{Full task results for fixed-commit-budget replay. Both selectors replay identical per-event commit counts, so their per-sample NFE matches.}
  \label{tab:revision_equal_budget_details}
  \begingroup
  \fontsize{9}{11}\selectfont
  \setlength{\tabcolsep}{3.5pt}
  \renewcommand{\arraystretch}{1.12}
  \providecolor{gaincolor}{RGB}{38,100,84}
  \providecommand{\revdelta}[1]{\textcolor{gaincolor}{#1}}
  \begin{tabular}{@{}llrrrrr@{}}
    \toprule
    Method & Metric & HumanEval & MBPP & GSM8K & MATH500 & Mean \\
    \midrule
    History-guided & ACC $\uparrow$ & 39.02 & 50.19 & 76.27 & 32.20 & 49.42 \\
    & NFE $\downarrow$ & 44.02 & 69.39 & 58.33 & 79.43 & 62.79 \\
    \addlinespace[2pt]
    Confidence-ranked & ACC $\uparrow$ & 34.76 & 38.52 & 67.55 & 23.80 & 41.16 \\
    & NFE $\downarrow$ & 44.02 & 69.39 & 58.33 & 79.43 & 62.79 \\
    \bottomrule
  \end{tabular}
  \endgroup
\end{table}

\subsection{Paired logit-reference interventions}
The Dream diagnostic uses the same 500 MATH500 prompts across four arms at seed 516,
temperature 0.2, top-$p$ 0.95, and generation length 256. The four arms use the
observed EMA reference, a frozen reference, no temporal support, and a
within-sample position-permuted reference. Confidence and the proposal-streak
gate remain fixed. Position permutation acts across positions, not across time.
The position curve aggregates all 256 generated slots into 16-position bins, with
paired prompt bootstrap intervals. The empirical distribution plots retain the
full prompt population and both tails. A trace change means a difference in a
position's (commit step, commit source) tuple; a text change means a difference
in the final generated string. All recorded writes across the four arms match
their local proposals.

\subsection{Sensitivity settings and History Gate windows}
\label{app:sensitivity}
Figure~\ref{fig:lambda_floor} evaluates seven settings on the same 164 HumanEval problems at generation length 256. The support-weight slice varies $\lambda$ with $w=0.35$ and $\tau_{\rm floor}=0.74$; the confidence-floor slice varies $\tau_{\rm floor}$ with $w=0.365$ and $\lambda=0.35$. Only the horizontal-axis parameter changes within each slice. Both use HG streak 3 and EMA decay 0.95. Each point pairs accuracy and NFE from the same complete run.

The sensitivity plot uses the same LLaDA HumanEval protocol as the main table and varies one control at a time. The system profiles below use matched requests to characterize the cost of different History Gate windows.

\section{Shared settings and system profiles}
\label{app:setting_ablation}
\subsection{Cross-task settings}
Table~\ref{tab:operating_settings_ablation} compares the benchmark-specific operating points with one vector shared across tasks and with the separate timing-study vector. The main text uses the benchmark-specific rows for the headline quality--NFE comparison.
\begin{table}[!htbp]
  \centering
  \caption{Effect of across-task and per-task parameter settings. Reference is Native Confidence; Mean averages the four complete tasks.}
  \label{tab:operating_settings_ablation}
  \begingroup
  \fontsize{9}{11}\selectfont
  \setlength{\tabcolsep}{2pt}
  \renewcommand{\arraystretch}{1.12}
  \providecolor{gaincolor}{RGB}{38,100,84}
  \providecommand{\revdelta}[1]{\textcolor{gaincolor}{#1}}
  \begin{tabular}{@{}lllrrrrr@{}}
    \toprule
    Method & Setting & Metric & HumanEval & MBPP & GSM8K & MATH500 & Mean \\
    \midrule
    \multicolumn{8}{@{}l}{\textit{LLaDA-8B}} \\
    Native Confidence & Reference & ACC $\uparrow$ & 37.20 & 47.08 & 77.10 & 30.60 & 48.00 \\
    &  & NFE $\downarrow$ & 52.28 & 83.34 & 74.26 & 96.48 & 76.59 \\
    \addlinespace[2pt]
    TACG & Across-task & ACC $\uparrow$ & 39.63 & 50.19 & 76.04 & 32.60 & 49.62 \\
    &  & NFE $\downarrow$ & 48.02 & 74.52 & 63.83 & 85.09 & 67.87 \\
    \addlinespace[2pt]
    TACG & Timing study & ACC $\uparrow$ & 42.07 & 50.19 & 77.03 & 32.60 & 50.47 \\
    &  & NFE $\downarrow$ & 45.71 & 69.35 & 67.12 & 85.09 & 66.82 \\
    \addlinespace[2pt]
    TACG & Default & ACC $\uparrow$ & 42.07 & 50.97 & 77.03 & 34.00 & 51.02 \\
    &  & NFE $\downarrow$ & 45.71 & 74.42 & 67.12 & 84.04 & 67.82 \\
    \addlinespace[3pt]
    \multicolumn{8}{@{}l}{\textit{Dream-7B}} \\
    Native Confidence & Reference & ACC $\uparrow$ & 47.56 & 63.42 & 73.16 & 39.00 & 55.79 \\
    &  & NFE $\downarrow$ & 51.19 & 82.46 & 74.63 & 94.60 & 75.72 \\
    \addlinespace[2pt]
    TACG & Across-task & ACC $\uparrow$ & 46.95 & 63.81 & 77.03 & 39.20 & 56.75 \\
    &  & NFE $\downarrow$ & 55.32 & 108.33 & 92.08 & 107.54 & 90.82 \\
    \addlinespace[2pt]
    TACG & Default & ACC $\uparrow$ & 50.00 & 63.04 & 73.62 & 39.60 & 56.56 \\
    &  & NFE $\downarrow$ & 50.80 & 93.53 & 74.26 & 91.14 & 77.43 \\
    \bottomrule
  \end{tabular}
  \endgroup
\end{table}

\subsection{Length and batch scaling}
\label{app:length_batch_scaling}
The scaling study uses a fixed 66-token prefix replicated across each batch on one H200 GPU (bfloat16, seed 516). It contains 16 conditions: four lengths for the original decoder at batch 1, and the same lengths for shared TACG at batches 1, 4, and 8. Each condition has five warm-ups, 20 timing repeats, and five separate memory repeats. The original decoder uses block length 32 and TACG uses 64. These profiles characterize throughput, latency, model-call count, and memory on the tested grid.

\begin{figure}[htbp]
  \centering
  \includegraphics[width=\linewidth]{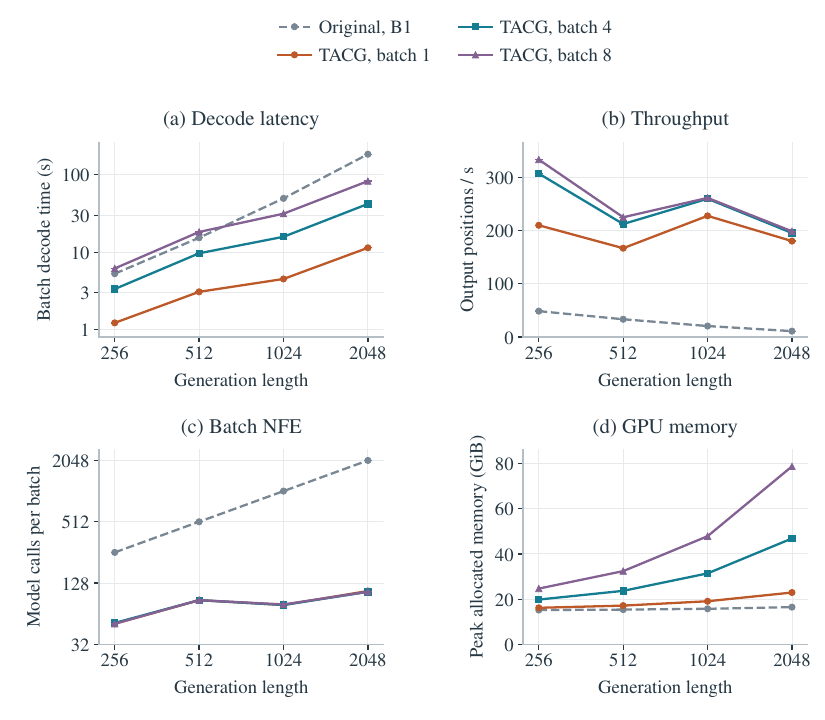}
  \caption{Length--batch scaling across four system metrics and 16 measured conditions on a single H200 with a fixed 66-token prefix replicated across batch. Curves connect measured lengths. Latency and throughput refer to decoding; NFE counts whole-batch model calls. Error bars are pointwise 95\% bootstrap intervals over repeated measurements of the same workload. Memory is the maximum allocated peak from five separate memory repeats. The original decoder uses block length 32; shared TACG uses 64.}
  \label{fig:length_batch_scaling}
\end{figure}

\begin{figure}[htbp]
  \centering
  \includegraphics[width=\linewidth]{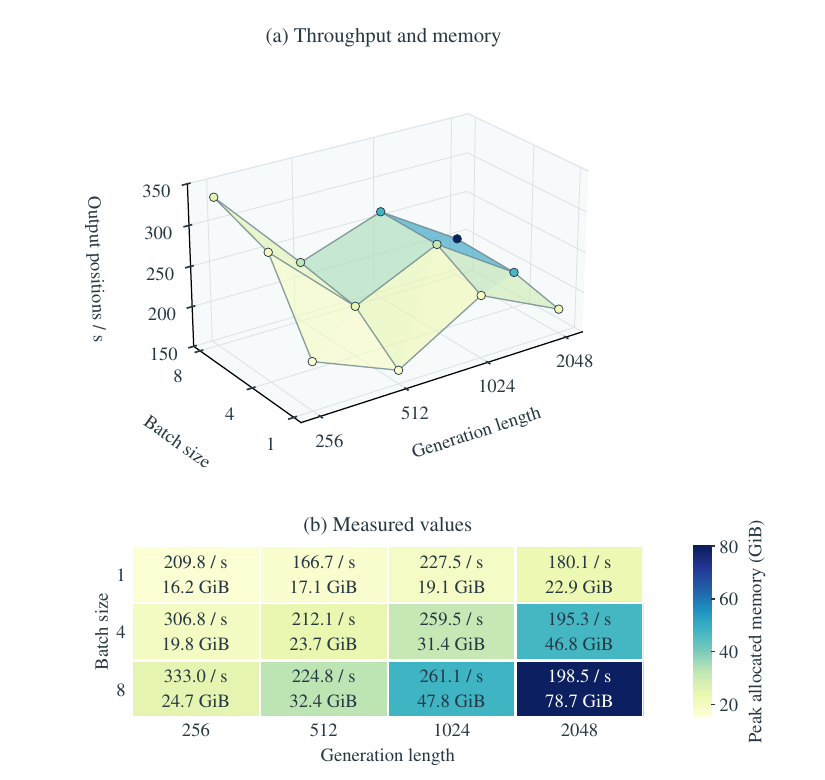}
  \caption{TACG throughput and memory on the measured length--batch grid. (a) Twelve nodes with length on a base-2 logarithmic axis. Translucent triangular faces connect measured vertices without adding observations; face colors encode their mean peak allocated memory. (b) All measured throughput and memory values. Throughput divides total generated output positions by total decode time; node/matrix colors show the largest allocated-memory peak across memory repeats.}
  \label{fig:throughput_memory_surface}
\end{figure}
The profiles show how generated length and batching interact with throughput and memory on this workload.

\section{Algebraic properties of temporal support}
\label{app:theory}

This appendix records algebraic identities for TACG's TILG readout. The derivations describe the implemented score and its role in commitment control; they do not assert posterior recovery or an optimal stopping rule.

\subsection{Temporal support is a historical log-ratio signal}

For derivation, write the auxiliary readout scores as
\begin{equation}
  u_t = (1+w)z_t - w\bar{z}_{t-1}.
\end{equation}
Let $p_t=\mathrm{softmax}(z_t)$, $p_t^{\mathrm{ref}}=\mathrm{softmax}(\bar{z}_{t-1})$, and $q_t=\mathrm{softmax}(u_t)$. Since logits are identifiable only up to an additive constant, we can write
\begin{align}
  \exp u_t(v)
  &= \exp\left((1+w)z_t(v)-w\bar{z}_{t-1}(v)\right) \\
  &\propto \frac{p_t(v)^{1+w}}{p_t^{\mathrm{ref}}(v)^w}.
\end{align}
After normalization,
\begin{equation}
  q_t(v) \propto p_t(v)\left(\frac{p_t(v)}{p_t^{\mathrm{ref}}(v)}\right)^w.
\end{equation}
The auxiliary readout therefore exposes a likelihood ratio between the current belief and the historical reference. Pairwise log-odds changes enter this readout explicitly. The bounded probability-gain score in Eq.~\ref{eq:supp-prob} additionally depends on normalization across the vocabulary.

\subsection{First-order belief-innovation interpretation}

Let $d_t=z_t-\bar{z}_{t-1}$ and $q_w=\mathrm{softmax}(z_t + w d_t)$. At $w=0$, $q_w=p_t$. The derivative of softmax gives
\begin{equation}
  \left.\frac{\partial q_w(v)}{\partial w}\right|_{w=0}
  = p_t(v)\left(d_t(v)-\mathbb{E}_{u\sim p_t}[d_t(u)]\right).
\end{equation}
Thus a small increase in $w$ raises the auxiliary readout value of tokens whose belief innovation is above the current average and lowers the readout value of tokens whose innovation is below that average. This supports the use of $q_t$ as a directional temporal signal and distinguishes it from confidence sharpening.

\subsection{Gauge invariance}

Categorical logits are not uniquely identified: adding a token-independent constant does not change the softmax. Suppose $z'_t(v)=z_t(v)+a$ and $\bar{z}'_{t-1}(v)=\bar{z}_{t-1}(v)+b$ for constants $a$ and $b$ that do not depend on token $v$. Then
\begin{equation}
  u'_t(v)=(1+w)z'_t(v)-w\bar{z}'_{t-1}(v)=u_t(v)+(1+w)a-wb.
\end{equation}
Normalization removes the added term because it is identical for every token, so $q_t$ remains unchanged. The temporal support statistic is therefore invariant to the additive gauge of logits at each position.

\subsection{Difference from KL-style stability signals}

Define the token-wise log-ratio
\begin{equation}
  r_t(v)=\log p_t(v)-\log p_t^{\mathrm{ref}}(v).
\end{equation}
A KL-style stability score aggregates this signal over the whole distribution,
\begin{equation}
  D_{\mathrm{KL}}(p_t\|p_t^{\mathrm{ref}})=\mathbb{E}_{v\sim p_t}[r_t(v)].
\end{equation}
This quantity measures distribution-level change. A candidate-level diagnostic of the same relative history is
\begin{equation}
  r_t(\hat{x}_i)=\log p_t(i,\hat{x}_i)-\log p_t^{\mathrm{ref}}(i,\hat{x}_i).
\end{equation}
The implemented probability gain $b_i$ also depends on the full normalization and is not generally a monotone function of this scalar log-ratio. The two forms of information answer different questions. KL-style stability asks whether the whole distribution is still changing. TILG supplies a signed candidate-conditioned contrast, while $b_i$ retains its positive gain for the readiness score. A distribution may change substantially while the current proposal is consistently reinforced, and a distribution may remain nearly unchanged while no candidate becomes more commit-ready. This distinguishes candidate-conditioned temporal support from distribution-level stability.

\end{document}